\documentclass[letterpaper, 10 pt, conference]{ieeeconf}  

\IEEEoverridecommandlockouts                              

\usepackage{graphicx} 
\usepackage{amsmath} 
\usepackage{amssymb}  
\usepackage{xcolor}
\usepackage{rotating} 
\usepackage{adjustbox} 
\usepackage{bm} 
\usepackage{multirow}
\usepackage{array} 

\usepackage[utf8]{inputenc}
\usepackage{url}
\usepackage{booktabs}
\usepackage{bbding}
\usepackage{pifont}
\usepackage{wasysym}
\usepackage{utfsym}
\usepackage{fontawesome}
\usepackage{epsfig} 
\usepackage{color}
\usepackage{times} 
\usepackage{mathptmx}
\usepackage{threeparttable}
\usepackage{booktabs}
\usepackage{hyperref}
\usepackage{graphicx}
\usepackage{amsmath}
\usepackage{booktabs}
\usepackage{makecell}
\usepackage{diagbox}
\usepackage{indentfirst}
\usepackage{titlesec}
\titlespacing*{\section}{0pt}{0.6\baselineskip}{0.3\baselineskip}  
\titlespacing*{\subsection}{0pt}{0.6\baselineskip}{0.3\baselineskip}  
\titlespacing*{\subsubsection}{0pt}{0.6\baselineskip}{0.4\baselineskip}  

\title{\LARGE \bf
JRN-Geo: A Joint Perception Network based on RGB and Normal images for Cross-view Geo-localization
}

\author{Hongyu Zhou$^{1}$, Yunzhou Zhang$^{1*}$, Tingsong Huang$^{2}$, Fawei Ge$^{1}$, Man Qi$^{1}$, Xichen Zhang$^{1}$, Yizhong Zhang$^{1}$
\thanks{*The corresponding author of this paper}
\thanks{$^{1}$Hongyu Zhou, Yunzhou Zhang, Fawei Ge, Man Qi, Xichen Zhang, and Yizhong Zhang are with College of Information Science and Engineering, Northeastern University, Shenyang
110819, China.
        {\tt\small zhangyunzhou@mail.neu.edu.cn}}%
\thanks{$^{2}$Tingsong Huang is with School of Computer Science, University of Sheffield, Sheffield S1 4DP, United Kingdom.}
\thanks{This project is funded by National Natural Science Foundation of China (No. 61973066), Major Science and Technology Projects of Liaoning Province(2021JH1/10400049).}%
}

\newcolumntype{P}{>{\centering\arraybackslash}p{1.2cm}}

\begin{document}

\maketitle
\thispagestyle{empty}
\pagestyle{empty}

\begin{abstract}
Cross-view geo-localization plays a critical role in Unmanned Aerial Vehicle (UAV) localization and navigation. However, significant challenges arise from the drastic viewpoint differences and appearance variations between images. Existing methods predominantly rely on semantic features from RGB images, often neglecting the importance of spatial structural information in capturing viewpoint-invariant features. To address this issue, we incorporate geometric structural information from normal images and introduce a Joint perception network to integrate RGB and Normal images (JRN-Geo). Our approach utilizes a dual-branch feature extraction framework, leveraging a Difference-Aware Fusion Module (DAFM) and Joint-Constrained Interaction Aggregation (JCIA) strategy to enable deep fusion and joint-constrained semantic and structural information representation. Furthermore, we propose a 3D geographic augmentation technique to generate potential viewpoint variation samples, enhancing the network’s ability to learn viewpoint-invariant features. Extensive experiments on the University-1652 and SUES-200 datasets validate the robustness of our method against complex viewpoint variations, achieving state-of-the-art performance.

\end{abstract}

\section{INTRODUCTION}
Cross-view geo-localization aims to match images of the same geographic location from different views, such as drone and satellite, and plays a crucial role in tasks like drone localization \cite{cvm} and navigation \cite{Dronet}. 
This technology has been widely applied across various fields, including aerial photography, precise delivery, and disaster detection. However, due to the visual content differences caused by viewpoint variations \cite{OriCNN}\cite{SAFA}\cite{NetVLAD} and global feature confusion \cite{CDFL}\cite{LFD}, cross-view geo-localization remains challenging. As a result, obtaining robust and discriminative feature representations has become a key focus in cross-view geo-localization research. 

Current cross-view geo-localization algorithms mainly focus on perceiving the detailed texture information of RGB images to construct descriptors. Some studies \cite{University-1652}\cite{LCM}\cite{DWDR} used full-image information to extract global features, but these features often became unstable when image semantics changed with viewpoint variations. Other methods \cite{LPN}\cite{FSRA}\cite{RK-Net} improved feature discriminability by extracting fine-grained local information, but fragmented contextual information in complex scenes led to feature misassociation. To address this, researchers \cite{MCCG}\cite{MFJR}\cite{MJRL} proposed multi-branch and multi-level interaction strategies to fuse global and local information, enabling joint representation of image features. However, these methods primarily considered RGB images, which were susceptible to viewpoint variations and relied mainly on prominent color and texture for discriminative cues, presenting inherent limitations.

\begin{figure}[t]
\centering
\includegraphics[width=\columnwidth]{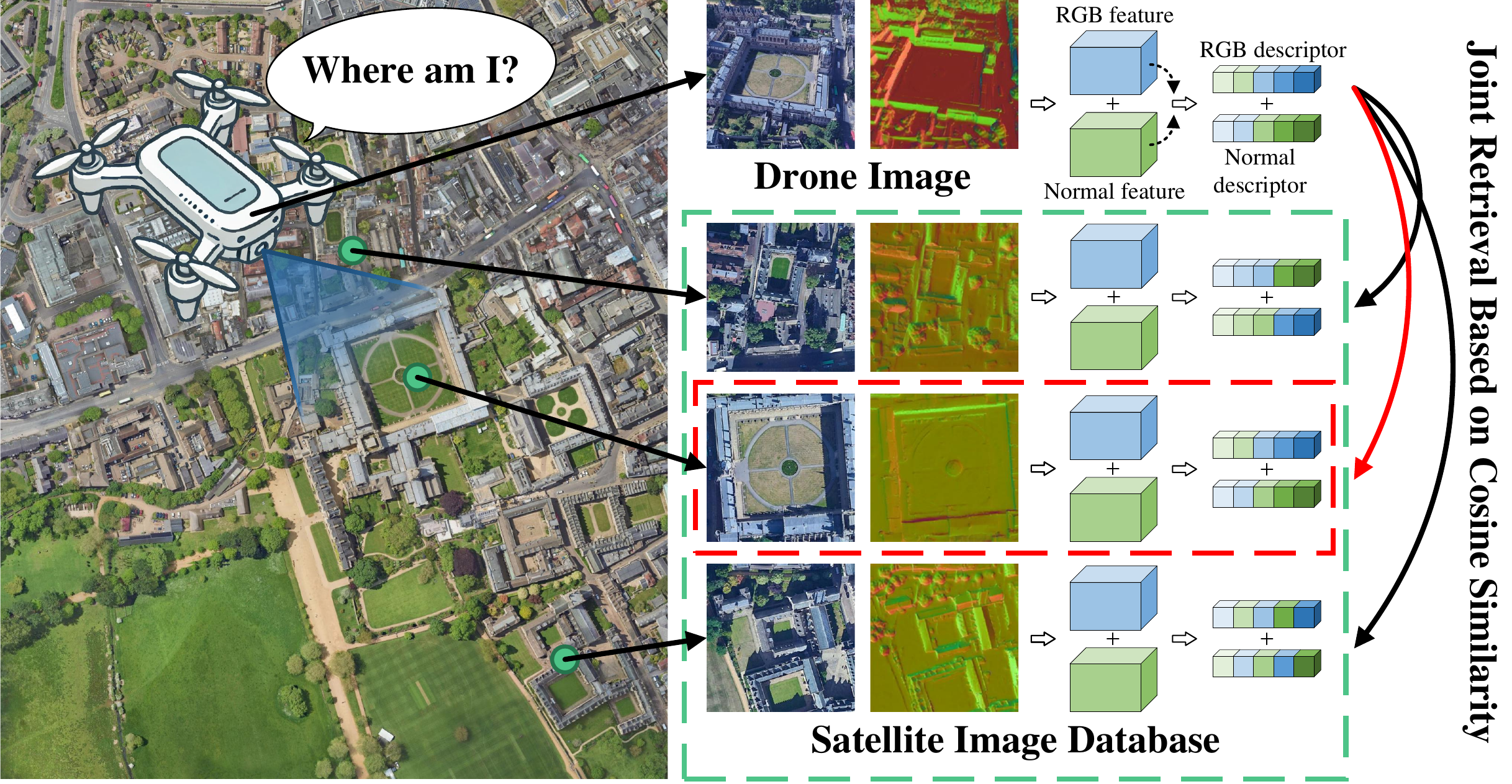}
\caption{\textbf{A cross-view geo-localization method based on joint perception of RGB and normal images.} The global descriptors are derived from RGB images captured by the camera and normal images generated through monocular normal estimation. Joint retrieval uses cosine similarity to measure the distance between drone and database descriptors.
}
\label{fig1}
\end{figure}

Compared to RGB images, normal images offered a stable representation of an object’s geometric structure, remaining robust against lighting and color variations. They could be easily generated using monocular normal estimation methods \cite{surface}\cite{GeoNet}\cite{Omnidata}. Some neural implicit surface reconstruction methods \cite{MonoSDF}\cite{VEGS} introduced normal geometry priors as additional supervision signals, significantly enhancing reconstruction quality and optimization efficiency. Normal images from aerial viewpoints contained 3D structural information of scenes, guiding the network to overcome the semantic information distribution shift caused by viewpoint variations and providing robust structural discriminative features. Building on these insights, we incorporate normal images into cross-view geo-localization and propose a network architecture that fuses RGB and normal image information interactively to mitigate the impact of viewpoint variations on localization accuracy, as illustrated in Fig.~\ref{fig1}.

Additionally, due to the difficulty in accurately annotating the geographic location at the center of drone images taken from oblique angles, obtaining matched pairs of drone and satellite data is challenging and incurs high costs. Existing cross-view geo-localization datasets \cite{University-1652}\cite{SUES-200} mostly used a geographic instance-based approach, collecting multiple drone images around building instances and obtaining one corresponding satellite image. This approach led to an imbalance in samples and limited the dataset size. To alleviate the impact of sample quality on algorithm accuracy, we propose a 3D geographic augmentation technique that first uses COLMAP \cite{colmap} to recover the spatial relationships between drone images and architectural scene, and then crops new cross-view instance data centered on marked spatial points from original images. This approach expands the sample size with varied viewpoints and helps the network learn viewpoint-invariant features.

Our contributions are listed as follows:
\begin{itemize}
\item We propose a novel and effective baseline, JRN-Geo, for cross-view geo-localization based on joint perception of RGB and normal images, designed to construct global descriptors robust to viewpoint variations.
\item We design a Difference-Aware Fusion Module (DAFM) to fuse semantic and structural information and a Joint-Constrained Interaction Aggregation (JCIA) strategy to aggregate multiple feature representations interactively.
\item A 3D geographic augmentation technique is designed to mine potential viewpoint variation samples without the necessity of additional annotations.
\item Extensive experiments on the University-1652 and SUES-200 datasets demonstrate that our model achieves state-of-the-art performance in cross-view geo-localization under complex viewpoint variations.
\end{itemize}

\section{Related Work}
\subsection{Cross-view Geo-localization}
The main challenges of cross-view geo-localization include complex viewpoint variations and sample imbalance. 
To alleviate viewpoint variations, PCL \cite{PCL} mitigated viewpoint variations by generating overhead views of drone images using a GAN network. LPN \cite{LPN} employed a ring partition strategy to guide the network’s focus on corresponding regions in cross-view images. MFJR \cite{MFJR} utilized part-level and patch-level features as feedback for global features to extract contextual information. 
In this paper, we tackle this challenge by leveraging the geometric information of normal images, capturing features that exhibit stronger robustness to viewpoint variations.

To address another challenge of sample imbalance and limited scene variety in datasets, FSRA \cite{FSRA} proposed a repeated sampling strategy to alleviate view sample imbalance. Sample4Geo \cite{sample4geo} used a custom sampler to train the model with drone-view sampling methods. Inspired by COLMAP \cite{colmap}, we propose a 3D geographic augmentation technique to deeply explore potential cross-view image pairs from existing data, effectively facilitating the network’s learning of viewpoint-invariant features.

\subsection{Geometry Information Fusion}
Geometry information fusion seeks to enhance feature representation by integrating geometric cues, such as surface normals or depth information, to improve the robustness of various vision tasks. 
Deepmvs \cite{Deepmvs} integrates multi-view depth information to predict high-quality disparity maps for 3D reconstruction.
MF-TGAA \cite{MF-TGAA} uses depth maps to provide structure awareness to VPR algorithms, overcoming the impact of dynamic environmental changes. 
In our method, we focus on combining RGB and normal image information. By using the DAFM and JCIA strategy, we capture and highly aggregate semantic and structural features, effectively improving the accuracy of cross-view geo-localization algorithms under complex viewpoint variations.

\section{Method}
\subsection{Overview} 
\begin{figure*}[t] 
\centering
\includegraphics[width=\textwidth]{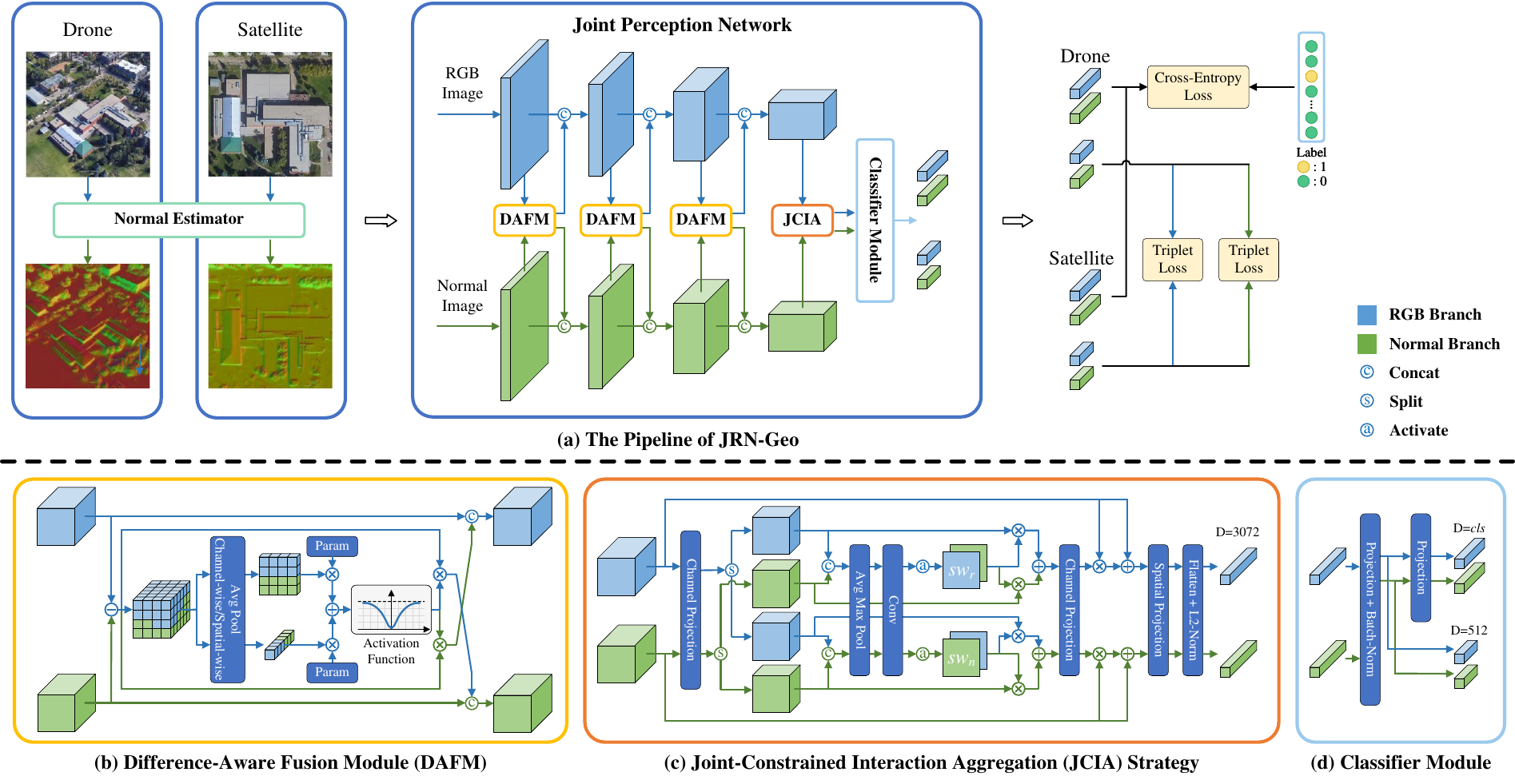}
\vspace{-17pt}
\caption{\textbf{Overview of our proposed framework for cross-view geo-localization. }(a) shows the pipeline of JRN-Geo, which integrates RGB and normal images for robust feature extraction. (b) DAFM focuses on the fusion of semantic and structural information. 
(c) JCIA strategy enhances spatial interaction and aggregates to obtain multi-feature representations. (d) The classifier module maps features for subsequent loss calculation.}
\label{fig2}
\vspace{-5pt}
\end{figure*}

We propose a cross-view geo-localization pipeline that combines RGB and normal information. First, a monocular normal estimation module generates normal images. Then, both RGB and normal images are fed into a dual-branch joint feature extraction network to extract features, using a DAFM to perceive and fuse differential features progressively. Next, the JCIA strategy achieves the combined representation of semantic and structural information. Finally, we find the most similar matching image by calculating the cosine similarity between the features of the retrieval image and the query image, and we use triplet loss and cross-entropy loss for supervised learning. The network structure is shown in Fig.~\ref{fig2}. Additionally, we designed a 3D geographic augmentation technique to mine potential cross-view image pairs, with the specific details described in Sections~\ref{sec:3D}.

\subsection{Monocular Normal Estimation} 
To obtain normal images, we use a surface normal estimation model from the Omnidata framework \cite{Omnidata}, which employs the DPT-Hybrid \cite{DPT} network for estimating pixel-level normal maps from the input images. DPT-Hybrid combines a ResNet50 backbone \cite{resnet} to compute image embeddings and Vision Transformer \cite{VIT} layers to capture long-range spatial dependencies, enabling the model to generate fine-grained pixel-level normal maps by progressively upsampling features at different resolutions. We chose this method for its strong generalization, achieved by training on large datasets like ScanNet \cite{Scannet} and NYU Depth V2 \cite{NYU}, making it suitable for use on datasets without ground-truth normals. In particular, we represent the image as \( x_{i} \in \mathbb{R}^{3 \times H \times W} \), where the subscript \( i \in \{r\, , \;n\} \) denotes the RGB and the corresponding normal images, respectively.

\subsection{Dual-Branch Feature Extraction
}

The color semantic information from RGB images and the geometric structural information from normal images can provide strong discriminative features for cross-view geo-localization tasks. However, the imbalanced image modalities make specific feature extraction challenging. Inspired by differential amplifier circuits that suppress common-mode signals and amplify differential-mode signals, we propose a dual-branch ConvNeXt \cite{convnext} network integrated with a Difference-Aware Fusion Module (DAFM), as illustrated in Fig.~\ref{fig2}(b). We first denote the intermediate features extracted from the \( m\)-th stage of the ConvNeXt network for RGB and normal images as \( f_{i}^m \) (\( m = 1\, , \; 2\, , \; 3 \)). Based on the principle of differential amplification circuits, \( f_{r}^m \) and \( f_{n}^m \) can be expressed in terms of their common-mode and differential-mode components as follows:
\begin{equation}
{f}_{r}^m = \frac{1}{2}({f}_{r}^m + {f}_{n}^m) 
+ \frac{1}{2}({f}_{r}^m - {f}_{n}^m),
\end{equation}
\begin{equation}
{f}_{n}^m = \frac{1}{2}({f}_{n}^m + {f}_{r}^m) 
+ \frac{1}{2}({f}_{n}^m - {f}_{r}^m),
\end{equation}
where the common-mode component reflects the combined features, while the differential-mode component captures the differential features in semantics and structure. The core idea of our DAFM is to capture these differential features through differential activation, progressively fusing semantic and structural information during the feature extraction process.

Specifically, we first subtract the intermediate features of RGB and normal images \( f_{r}^m \, , \; f_{n}^m \), to obtain the differential feature \( f_{d}^m \). We then apply spatial average pooling \( SP(\cdot) \) and channel average pooling \( CP(\cdot) \) to obtain statistical weights and use learnable parameters \( p_s \) and \( p_c \) to learn inductive biases. 
To suppress common-mode signals and amplify differential-mode signals, we process the derivative of the sigmoid function \( sigmoid’(\cdot) \) by flipping and scaling it to obtain an activation function \( \delta(\cdot) \):
\begin{equation}
\delta(x) = 1 - 4 \times sigmoid’(x),
\end{equation}
where \( \delta(\cdot) \) is an even function, which helps eliminate the influence of the positive and negative values of \( f_{d}^m \) and compress weights into the \((0\, , \;1)\) range.
We multiply the activated weights by \( f_{d}^m \) and then concatenate the result with the original features \( f_{r}^m \, , \; f_{n}^m \) along the channel dimension, generating the joint features \( \tilde{f}_{r}^m \, , \; \tilde{f}_{n}^m \) that contain significant differential information, which serves as inputs for the next stage. The above process can be precisely described as follows:
\begin{align}
\tilde{f}_{r}^m &= concat((f_{r}^m, (f_{r}^m - f_{n}^m) \odot \delta\left(p_s \odot SP(f_{r}^m - p_{n}^m) \right) \notag \\ 
&\quad \oplus \left(p_c \odot CP(f_{r}^m - f_{n}^m)\right)),
\label{eq4}
\end{align}
where \( \odot \) denotes element-wise multiplication. Equation~(\ref{eq4}) shows the computation process for the RGB branch, with the same applies to the normal branch.
The final feature output by the dual-branch network \( F_{dual} \) can be represented as:
\begin{equation}
{f}_{{r}}\, , \; {f}_{{n}} = F_{dual}({x}_{{r}}\, , \; {x}_{{n}}),
\end{equation}
\noindent where \( f_{r}\, , \; f_{n} \in \mathbb{R}^{C \times H \times W} \) represent the final output features of RGB and normal images.

\subsection{Joint-Constrained Interaction Aggregation Strategy}
To combine the intensity of semantic and structural spatial perception and obtain discriminative descriptors, we propose a Joint-Constrained Interaction Aggregation (JCIA) strategy to interact with dual-branch information and output constrained aggregated features based on each branch. The detailed structure is shown in Fig.~\ref{fig2}(c).

Firstly, we perform a channel-wise linear mapping on \( f_{r}\, , \; f_{n} \) and then split them channel-wise by averaging to obtain the split features \( {f’}_{r}^q\, , \; {f’}_{n}^q \in \mathbb{R}^{C/2 \times H \times W} \) (\( q = 1\, , \; 2 \)). The features \( \hat{f}_{r}^q \, , \; \hat{f}_{n}^q \) are cross-concatenated to allow interaction between RGB and normal features. Next, channel-wise average pooling and max pooling operations \( P(\cdot) \) are applied to extract spatial relationships, followed by convolution \( C^{2 \rightarrow 2}(\cdot) \) and the sigmoid activation function \( s(\cdot) \) to obtain the cross-branch spatial interaction weights \( SW_r \, , \; SW_n \) for each branch:
\begin{equation}
{SW}_{{r}}\, ; \;{S}{W}_{{n}} = s(C^{2 \rightarrow 2}(P([{{f}’}_{{r}}^{1}, {{f}’}_{{n}}^{1}; {{f}’}_{{r}}^{2}, {{f}’}_{{n}}^2]))),
\end{equation}
where \( SW_r\, , \, SW_n \in \mathbb{R}^{2 \times H \times W} \). We separate \( SW_r\, , \; SW_n \) channel-wise to obtain \( {SW}_r^q \, , \, {SW}_n^q \in \mathbb{R}^{1 \times H \times W} \) (\( q \in \{1,2\} \)), and multiply them with \( {f’}_{r}^q \, , \, {f’}_{n}^q \) then add, applying channel-wise projection \( F_c^{C/2 \rightarrow C}(\cdot) \) to obtain global interactive attention weights \( GW_r \, , \, GW_n \):
\begin{equation}
{G}{W}_{{r}} = P_c^{C/2 \rightarrow C} ({{SW}}_{{r}}^{1} \times {{f}’}_{{r}}^{1} + {{SW}}_{{r}}^{2} \times {{f}’}_{{n}}^{1}),
\end{equation}
\begin{equation}
{G}{W}_{{n}} = F_c^{C/2 \rightarrow C} ({{SW}}_{{n}}^{1} \times {{f}’}_{{r}}^{2} + {{SW}}_{{n}}^{2} \times {{f}’}_{{n}}^{2}),
\end{equation}
where \( GW_r, GW_n \in \mathbb{R}^{D \times H \times W} \). 
We multiply \( GW_r \, , \, GW_n \) with the input feature maps \( f_{i,r} \, , \; f_{i,n} \) element-wise and apply residual connections to obtain the dual-branch interactive feature maps \( \overline{f_r} \, , \; \overline{f_n}  \in \mathbb{R}^{C \times H \times W}\). The resulting feature maps are flattened along the spatial dimensions and aggregated using a linear mapping \( F_s^{H \times W \rightarrow d}(\cdot) \), followed by channel-wise flattening and the application of L2 normalization \( Norm_{L_2}(\cdot) \) to extract the RGB and normal joint-constrained interaction aggregation vectors \( g_{i,r} \, , \, g_{i,n} \):
\begin{equation}
{g}_{{r}}\, ; \;{g}_{{n}} = Norm_{L2} \ (Flatten(F_s^{H\times W\rightarrow d}(\overline{f_r}\, ; \;\overline{f_n}))),
\end{equation}
where \( g_{r} \, , \; g_{n} \in \mathbb{R}^{d \times c} \), with \( d \) representing the spatial mapping dimensions, set to \(d=3\) in our experiments. 

\subsection{3D Geographic Augmentation Technique}
\label{sec:3D}
\begin{figure}[t]
\centering
\includegraphics[scale=0.525]{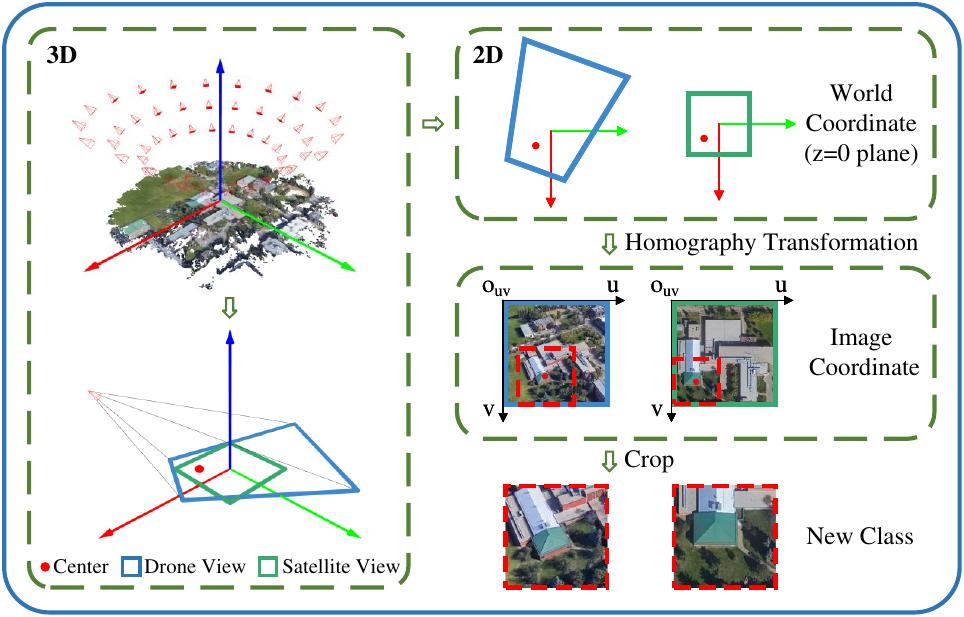}
\caption{\textbf{The 3D geographic augmentation technique. }The left side shows the projection of camera imaging regions in 3D space, while the right side illustrates 2D homography transformation and cropping.
}
\label{fig3}
\end{figure}

To deeply leverage the cross-view viewpoint variation samples in the data, we propose a 3D geographic augmentation technique. This approach aims to enhance sample diversity by mining potential cross-view instance pairs, facilitating the network to learn viewpoint-invariant features thoroughly. The detailed process is shown in Fig.~\ref{fig3}. 

Firstly, we use COLMAP \cite{colmap} for 3D point cloud reconstruction and pose estimation from drone image sequences. 
To align the coordinate system’s plane \( P_{XOY} \) approximately parallel to the ground of the point cloud, we apply simple rotation and translation adjustments on the point cloud and camera poses in 3D space. 
The spatial point \( c = (x_c\, , \,y_c\, , \,0) \) on \( P_{XOY} \) is designated as the new geographic instance center, and a neighborhood \( S(c) \) centered at \( c \) is defined:
\begin{equation}
\small 
S(c) = \{(x, y, z) \ | \ x \in [x_c - r, x_c + r], \ y \in [y_c - r, y_c + r] \},
\end{equation}
where \( r \) is the grid size of the neighborhood. We calculate the mean Z-coordinate \(\bar{z}\) of all points within \( S(c) \) as the height offset and translate the point cloud and camera poses along the Z-axis by \(-\bar{z}\) distance. This adjustment makes the points in \( S(c) \) close to \( P_{XOY} \), ensuring that point \( c \) is near the building surface
Next, based on the camera’s intrinsic parameters and pose, we calculate the intersection of the camera’s frustum rays with \( P_{XOY} \), forming a trapezoidal area \( r_d \) that represents the drone’s imaging region. Since the satellite and drone images of the geographic instance have relatively fixed shooting orientations, we manually annotate the satellite image’s trapezoidal imaging region \( r_s \) on \( P_{XOY} \). At this point, the imaging region of both the drone and satellite images concerning point \( c \) are aligned with the 2D plane \( P_{XOY} \). 
We then calculate the homography matrices \( H_d\, , \; H_s \) to project \( r_d \, , \; r_s \) into the pixel coordinate systems of the drone and satellite images. Using these matrices, we map point \( c \) to the image pixel coordinates to obtain the mapped points \( \widetilde{c}_d \, , \; \widetilde{c}_s \):
\begin{equation}
\widetilde{{c}}_d = {H}_d , \quad \widetilde{{c}}_s = {H}_s \mathbf{c},
\end{equation}
where matrices \( H_d \) and \( H_s \) can be calculated from the corresponding vertex coordinates of the projection regions and the pixel coordinate system. Details of the calculation are omitted here. We compute the shortest distances \(d\) from the projection points to the top, bottom, left, and right edges of the image and set thresholds \( d_{max} \) and \( d_{min} \) to limit the cropping size \( d_{cut} \):
\begin{equation}
d_{cut} = 
\begin{cases}
d_{max}, & d \in [d_{max}\, , \; +\infty), \\
d, & d \in [d_{min}\, , \;d_{max}), \\
None, & d \in [0\, , \;d_{min}),
\end{cases}
\end{equation}
when \( d \in [d_{max}\, , \;+\infty) \), the image overlap is too high, so we set \( d_{cut} = d_{max} \); when \( d \in [0\, , \;d_{min}) \), the image region is too small, and cropping is not applied. 
We defined nine candidate center positions based on image proportions and set an augmentation factor \( k \). We randomly select \( k \) spatial centers and perform batch cropping to generate new geographic instance data, enhancing sample diversity with viewpoint variations. Ablation studies in Section~\ref{Ablation} confirm that accuracy is optimal with \( k = 4 \).

\subsection{Loss Function}
We employ a simple classifier module to further map the interaction aggregation vectors \( g_{r} \, , \; g_{n} \) and set optimization objectives for supervised learning, as shown in Fig.~\ref{fig2}(d). The intermediate vectors \( v_{r} \, , \; v_{n} \in \mathbb{R}^{512}\) originating from the same branch but different views, are used to compute the triplet loss \( L_{Triplet} \), which aims to minimize intra-class distances and maximize inter-class distances among the encoded vectors. Simultaneously, the classification vectors \( z_{r} \, , \; z_{n} \in \mathbb{R}^{cls}\) (where \(cls\) denotes the number of classes) are used to calculate the cross-entropy loss \( L_{CrossEntropy} \), which refines the overall distribution of the instance data and guides the optimization direction of the triplet loss. The total loss is expressed as:
\begin{equation}
L_{total} = L_{Triplet} + L_{CrossEntropy}.
\end{equation}

\begin{table}[t]
\renewcommand{\arraystretch}{1}
    \centering
    \caption{Comparison with state-of-the-art methods on the University-1652 dataset. The best and second-best results are highlighted in red and blue, respectively.}
    \label{table_example1}
    \begin{adjustbox}{max width=\columnwidth}
    \normalsize  
    \begin{tabular}{cccccc}
        \Xhline{1pt}  
        \multirow{2}{*}{Method} & \multirow{2}{*}{Publication} 
        & \multicolumn{2}{c}{Drone $\rightarrow$ Satellite} 
        & \multicolumn{2}{c}{Satellite $\rightarrow$ Drone} \\
        & & R@1 & AP & R@1 & AP \\
        \hline
        LPN \cite{LPN} & TCSVT’22 & 75.93 & 79.14 & 86.45 & 74.79 \\
        FSRA \cite{FSRA} & TCSVT’22 & 85.50 & 87.53 & 89.73 & 84.94 \\
        MBF \cite{mbf} & Sensors’23 & 89.05 & 90.61 & 93.15 & 88.17 \\
        MCCG \cite{MCCG} & TCSVT’23 & 89.64 & 91.32 & 94.30 & 89.39 \\
        Sample4Geo \cite{sample4geo} & ICCV’23 & 92.65 & 93.81 & 95.14 & 91.39 \\
        MFJR \cite{MFJR} & TGRS’24 & 91.87 & 93.15 & 95.29 & 91.51 \\
        Ours(k=0) & \textemdash & \textcolor{blue}{94.32} & \textcolor{blue}{95.29} & \textcolor{blue}{96.15} & \textcolor{blue}{93.81} \\
        Ours(k=4) & \textemdash & \textcolor{red}{95.13} & \textcolor{red}{95.85} & \textcolor{red}{96.72} & \textcolor{red}{94.93} \\
        \Xhline{1pt}  
    \end{tabular}
    \end{adjustbox}
    \vspace{10pt} 
\end{table}

\section{Experiment}
\subsection{Experimental Preparation}
\textbf{Datasets}.
We evaluate our method on two public geo-localization datasets: University-1652 \cite{University-1652}, a large-scale dataset for drone geo-localization, and SUES-200 \cite{SUES-200}, which includes drone images captured at multiple heights.

\textbf{Evaluation Metrics.}
We use Recall@K (R@K) and Average Precision (AP) to evaluate model performance. R@K measures the proportion of correctly matched images in the top K results, indicating retrieval accuracy, while AP calculates the area under the Precision-Recall curve to consider both precision and recall.

\textbf{Implementation Details.}
Our experiments use a ConvNeXt-Base backbone pre-trained on ImageNet with a learning rate of 0.001. For feature extraction, we employ separate backbones with independent weights for the RGB and normal branches, while sharing the same network weights for both drone and satellite views. The input RGB and normal images are resized to 384×384 pixels. During the testing phase, cosine similarity is used to measure the similarity between the query and candidate images.

\subsection{Comparison with the State-of-the-art Methods}

\textbf{Results on the University-1652 Dataset.}
We evaluated our method on two tasks: Drone Localization (Drone → Satellite) and Drone Navigation (Satellite → Drone). As shown in Table~\ref{table_example1}, our joint perception approach using RGB and normal images outperforms others, even without instance augmentation. With the augmentation technique, the adequate viewpoint variation samples further improve R@1 and AP from the initial 95\%, demonstrating the robustness and effectiveness of our approach. Compared to the MBF  \cite{mbf}, which is a multi-modal approach combining text and image information, our approach achieves an average 5\% improvement across all four metrics, validating the advantages of the structural information from normal maps. The results indicate that our method excels in environments with viewpoint variations, effectively integrating semantic and structural information to overcome the limitations of single-modality RGB images and leveraging existing data to enhance learning of viewpoint-invariant features.

\begin{table}[t]
    \centering
    \tabcolsep=0.13cm
    \caption{Evaluating Viewpoint Changes on the SUES-200 Dataset. The best and second-best results are highlighted in red and blue, respectively.}
    \label{table_example2}
    \begin{adjustbox}{max width=\columnwidth}
    \normalsize
    \begin{tabular}{ccccccccc}
        \Xhline{1pt}
        \multicolumn{9}{c}{Drone → Satellite }\\
        Method & \multicolumn{2}{c}{150m} & \multicolumn{2}{c}{200m} & \multicolumn{2}{c}{250m} & \multicolumn{2}{c}{300m} \\ 
        & R@1 & AP & R@1 & AP & R@1 & AP & R@1 & AP \\ 
        \hline
        Baseline \cite{University-1652} & 55.65 & 61.92 & 66.78 & 71.55 & 72.00 & 76.43 & 74.05 & 78.26 \\
        LPN \cite{LPN} & 61.58 & 67.23 & 70.85 & 75.96 & 80.38 & 83.80 & 81.47 & 84.53 \\
        MBF \cite{mbf} & 85.62 & 88.21 & 87.43 & 90.02 & 90.65 & 92.53 & 92.12 & 93.63 \\ 
        MCCG \cite{MCCG} & 82.22 & 85.47 & 89.30 & 91.41 & 93.82 & 95.04 & 95.07 & 96.20\\ 
        MFJR \cite{MFJR} & \textcolor{blue}{88.95} & \textcolor{blue}{91.05} & 93.60 & 94.72 & 95.42 & 96.28 & 94.45 & \textcolor{blue}{97.84} \\ 
        Ours(k=0) & 86.15 & 89.02 & \textcolor{blue}{93.80} & \textcolor{blue}{97.12} & \textcolor{blue}{97.12} & \textcolor{blue}{97.77} & \textcolor{blue}{96.22} & 97.10 \\ 
        Ours(k=4) & \textcolor{red}{96.47} & \textcolor{red}{97.26} & \textcolor{red}{98.60} & \textcolor{red}{98.92} & \textcolor{red}{99.28} & \textcolor{red}{99.45} & \textcolor{red}{99.10} & \textcolor{red}{99.33} \\ 
        \hline
        \multicolumn{9}{c}{Satellite → Drone}\\ 
        Method & \multicolumn{2}{c}{150m }& \multicolumn{2}{c}{200m}& \multicolumn{2}{c}{250m}& \multicolumn{2}{c}{300m }\\ 
        ~ & R@1 & AP & R@1 & AP & R@1 & AP & R@1 & AP \\ 
        \hline
        Baseline \cite{University-1652} & 55.65 & 61.92 & 66.78 & 71.55 & 72.00 & 76.43 & 74.05 & 78.26 \\ 
        LPN \cite{LPN} & 83.75 & 66.78 & 88.75 & 75.01 & 92.50 & 81.34 & 92.50 & 85.72 \\ 
        MBF \cite{mbf} & 88.75 & 84.74 & 91.25 & 89.95 & 93.75 & 90.65 & \textcolor{blue}{96.25} & 91.60\\ 
        MCCG \cite{MCCG} & 93.75 & 89.72 & 93.75 & 92.21 &96.25 & 96.14 & 95.00 & 92.03 \\ 
        MFJR \cite{MFJR} & 95.00 & 89.31 & \textcolor{blue}{96.25} & \textcolor{blue}{94.69} & \textcolor{blue}{97.50} & \textcolor{blue}{96.92} & \textcolor{red}{98.75} & 97.14 \\ 
        Ours(k=0) & \textcolor{blue}{97.50} & \textcolor{blue}{91.09} & \textcolor{red}{98.75} & 93.85 & \textcolor{red}{98.75} & 96.77 & \textcolor{red}{98.75} & \textcolor{blue}{97.92} \\ 
        Ours(k=4) & \textcolor{red}{98.75} & \textcolor{red}{96.05} & \textcolor{red}{98.75} & \textcolor{red}{98.02} & \textcolor{red}{98.75} & \textcolor{red}{99.06} & \textcolor{red}{98.75} & \textcolor{red}{98.97} \\ 
        \Xhline{1pt}
    \end{tabular}
    \end{adjustbox}
\end{table}

\begin{table}[h]
    \centering
    \caption{Study on Effectiveness of Normal Information On University-1652. }
    \label{table3}
    \begin{tabular}{ccccc}
        \Xhline{1pt}  
        \multirow{2}{*}{Input} & \multicolumn{2}{c}{Drone → Satellite }& \multicolumn{2}{c}{Satellite → Drone }\\ 
        & R@1 & AP & R@1 & AP \\ \hline
        Normal & 37.33 & 43.13 & 58.63 & 51.10\\ 
        RGB & 88.98 & 90.34 & 91.87 & 87.86 \\ 
        RGB+Normal & \textbf{91.58} & \textbf{92.46} & \textbf{94.44} & \textbf{91.51} \\ 
        \Xhline{1pt}  
    \end{tabular}
\end{table}

\textbf{Results on the SUES-200 Dataset.}
To evaluate the robustness of our method under varying levels of viewpoint variation, we conducted experiments on the SUES-200 dataset at altitudes of 150m, 200m, 250m, and 300m. The results, shown in Table~\ref{table_example2}, indicate that our method consistently achieves optimal or near-optimal performance across all tested altitudes. Notably, at the low altitude of 150m, our 3D geographic augmentation technique significantly enhances the network’s ability to learn viewpoint-invariant features by fully utilizing potential viewpoint variation samples, resulting in high localization accuracy.
Further analysis of the R@1 and AP across different altitudes reveals that the method effectively captures viewpoint-invariant features by leveraging the structural information of normal images. These results indicate that the proposed approach maintains strong robustness regardless of altitude, performing well under severe viewpoint differences at low altitudes and minor differences at higher altitudes.
\begin{table}[t]
    \centering
    \caption{Ablation of Methods Effectiveness on University-1652 Dataset
    }
    \label{table4}
    \begin{tabular}{cccccc}
        \Xhline{1pt}  
        \multirow{2}{*}{DAFM} & \multirow{2}{*}{JCIA} & 
        \multicolumn{2}{c}{Drone → Satellite }& \multicolumn{2}{c}{Satellite → Drone}\\ 
        & & R@1 & AP & R@1 & AP \\ \hline
        $\times$& $\times$ & 91.58 & 92.46 & 94.44 & 91.51 \\
        $\checkmark$ & $\times$ & 92.69 & 93.68 & 95.15 & 92.83 \\ 
        $\times$ & $\checkmark$ & 93.47 & 94.32 & 95.58 & 93.23 \\ 
        $\checkmark$ & $\checkmark$ & \textbf{94.32} & \textbf{95.29} & \textbf{96.15} & \textbf{93.81} \\    \Xhline{1pt}  
    \end{tabular}
\end{table}
\begin{figure}[t]
    \centering
    \includegraphics[scale=0.5]{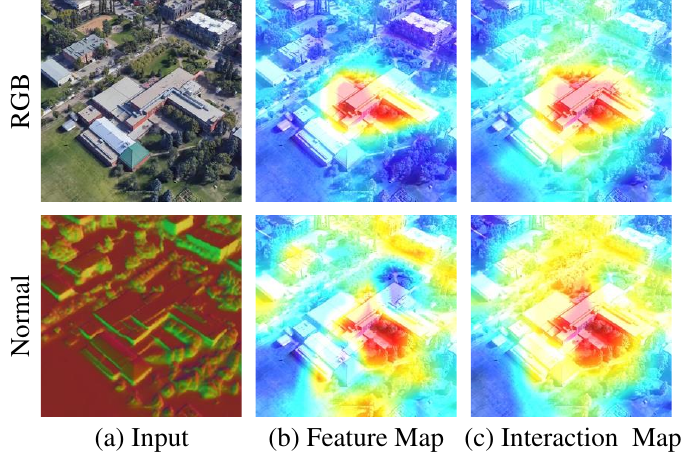}
    \vspace{-3pt}
    \caption{\textbf{Visualization of RGB and normal feature heatmaps.} (a) shows the input Images. (b) shows that RGB focuses on the main building information, while normals specialize in capturing the global structure. (c) demonstrates the post-interaction feature maps integrating the spatial attention advantages of both modalities.
    }
    \label{fig4}
\end{figure}
\subsection{Ablation Studies}
\label{Ablation}
\textbf{Evaluating the Effectiveness of Normal Information.}
To validate the effectiveness of normal information in cross-view geo-localization tasks, we removed all modules to avoid dual-branch feature fusion operations and used only the ConvNet backbone network to extract image features without augmentation. 
We tested different image combinations on the University-1652 dataset, as shown in Table~\ref{table3}. From the table, it can be observed that the network performs poorly with only normal images, while the best performance is achieved when both RGB and normal images are used, with R@1 and AP improving by an average of 1.7\% compared to using RGB images alone. 
The experimental results show that normal images lack detailed texture information of buildings and cannot independently support cross-view image retrieval. However, their geometric information describes the scene’s structural distribution, and when combined with color semantics, it enables multi-dimensional joint retrieval, leading to optimal performance. Experiments show that incorporating normal information significantly improves cross-view geo-localization accuracy.

\textbf{Ablation Study on the Proposed Components.}
To verify the effectiveness of the proposed DAFM and JCIA strategy, we conducted ablation experiments without augmentation and visualized heatmaps on the University-1652 dataset. As shown in Table~\ref{table4}, using either DAFM or JCIA alone improves R@1 and AP compared to not using any fusion methods, with the best results achieved when both are combined. Heatmap visualizations in Fig.~\ref{fig4}(b) show that the DAFM helps the dual-branch network focus on salient semantic and structural regions, while the JCIA effectively integrates spatial attention between RGB and normal features as shown in Fig.~\ref{fig4}(c). These findings demonstrate that DAFM and JCIA enable the network to jointly perceive RGB and normal features, capturing robust discriminative descriptors resistant to viewpoint variations, confirming the effectiveness of the proposed components.

\begin{figure}[t]
    \centering
        \vspace{3pt}
    \includegraphics[width=\columnwidth]{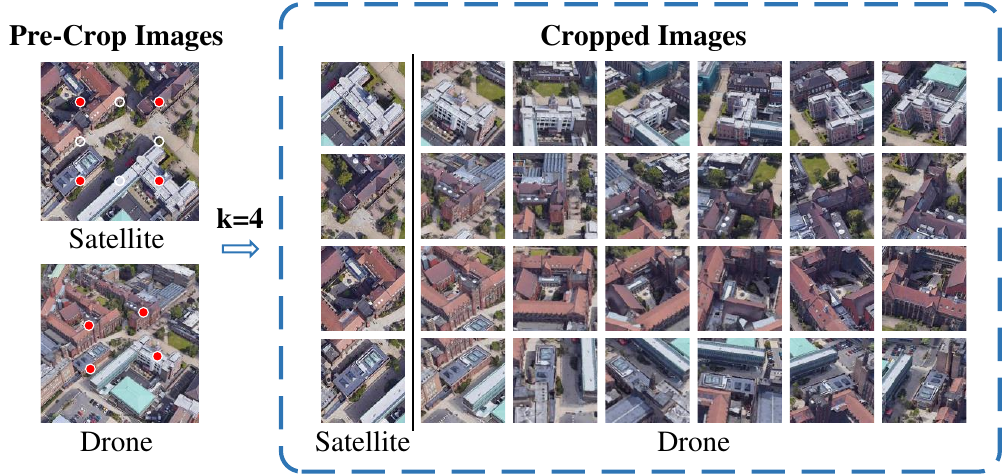}
    \caption{\textbf{Visualization of Augmentation Images.}
    The left shows images before cropping and the right shows the augmented instance images with the augmentation factor \(k=4\). All augmented images are centered on the marked spatial points.}
    \label{fig5}
\end{figure}

\begin{figure}[t]
    \centering
    \includegraphics[width=\columnwidth]{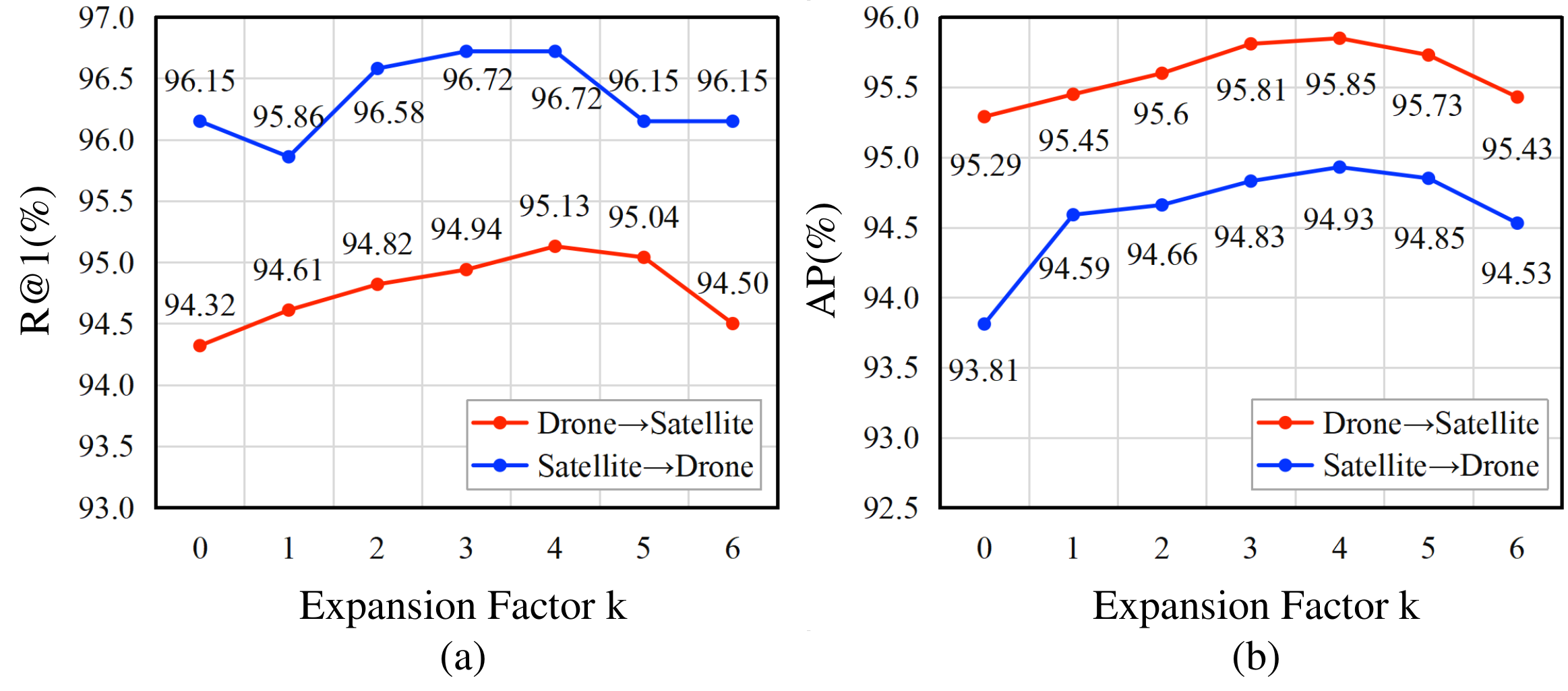}
    \vspace{-17pt}
    \caption{\textbf{Comparison of the effects of the augmentation factor \(k\). }(a) shows the impact of the augmentation factor \(k\) on Recall@1, while (b) shows its impact on AP. Both Recall@1 and AP achieve their optimal performance when \(k=4\).}
    \label{fig6}
\end{figure}

\textbf{Impact of Augmentation Factor \(k\).}
To investigate how the augmentation data affects the algorithm’s accuracy, we presented the augmented images and then conducted experiments with different number of augmentation factor \(k\) on the University-1652 dataset. As shown in Fig.~\ref{fig5}, the cropped images are centered on the marked points, serving as high-quality cross-view samples that expand the dataset effectively. From Fig.~\ref{fig6}, it can be observed that as \(k\) increases from 1 to 4, both AP and R@1 continue to improve, reaching overall optimal performance when \(k=4\). However, as \(k\) increases further, the algorithm’s accuracy plateaus or may even decline. The experimental results demonstrate that our method effectively identifies more potential cross-view sample pairs, significantly enhancing algorithm accuracy. Nevertheless, as \(k\) increases beyond 4, the overlap of new instances grows, confusing the network’s ability to distinguish between features of different instances. Therefore, selecting an appropriate augmentation factor is crucial for the algorithm’s performance.

\section{CONCLUSIONS}
In this paper, we propose a novel cross-view geo-localization algorithm designed to mitigate challenges posed by viewpoint variations. By incorporating structural information from normal images and developing a dual-branch feature extraction framework, our approach effectively captures both RGB and geometric features. The integration of joint constraints helps to avoid feature confusion while ensuring the extraction of multiple discriminative features. Furthermore, the proposed 3D geographic augmentation technique expands the dataset with potential viewpoint variation samples, enhancing the network’s ability to learn viewpoint-invariant features. Extensive experiments on the University-1652 and SUES-200 datasets demonstrate that our method achieves state-of-the-art performance under complex conditions. Looking forward, we aim to explore the potential of applying multi-image information in dynamic scenes and handling multi-scale variations to further improve the accuracy of cross-view geo-localization.






\bibliographystyle{ieeetr}
\bibliography{references}

\end{document}